\documentclass[11pt]{amsart}

\usepackage[letterpaper, margin=1in]{geometry}
\usepackage{amsmath, amssymb, amsthm, amsaddr}
\usepackage{enumerate, subcaption, graphicx, hyperref}
\usepackage{natbib}
\date{\today} 

\title{Event detection from novel data sources: Leveraging satellite imagery alongside GPS traces}

\author{Ekin Ugurel}
\address{University of Washington, Seattle, WA
\\ \texttt{ugurel@uw.edu}}

\author{Steffen Coenen}
\address{DKS Associates, Seattle, WA\\
\texttt{scoenen@uw.edu}}

\author{Minda Zhou Chen}
\address{University of Washington, Seattle, WA\\
\texttt{mindac@uw.edu}}

\author{Cynthia Chen}
\address{University of Washington, Seattle, WA\\ \texttt{qzchen@uw.edu}}

\begin{document}



\maketitle

\section{Abstract}

Rapid identification and response to breaking events, particularly those that pose a threat to human life such as natural disasters or conflicts, is of paramount importance.  The prevalence of mobile devices and the ubiquity of network connectivity has generated a massive amount of temporally- and spatially-stamped data. Numerous studies have used mobile data to derive individual human mobility patterns for various applications. Similarly, the increasing number of orbital satellites has made it easier to gather high-resolution images capturing a snapshot of a geographical area in sub-daily temporal frequency. We propose a novel data fusion methodology integrating satellite imagery with privacy-enhanced mobile data to augment the event inference task, whether in real-time or historical. In the absence of boots on the ground, mobile data is able to give an approximation of human mobility, proximity to one another, and the built environment. On the other hand, satellite imagery can provide visual information on physical changes to the built and natural environment. The expected use cases for our methodology include small-scale disaster detection (i.e., tornadoes, wildfires, and floods) in rural regions, search and rescue operation augmentation for lost hikers in remote wilderness areas, and identification of active conflict areas and population displacement in war-torn states. Our implementation is open-source on GitHub: \url{https://github.com/ekinugurel/SatMobFusion}.

\newpage

\section{Introduction}
\begin{figure}          \centering
\includegraphics[width=0.7\textwidth]{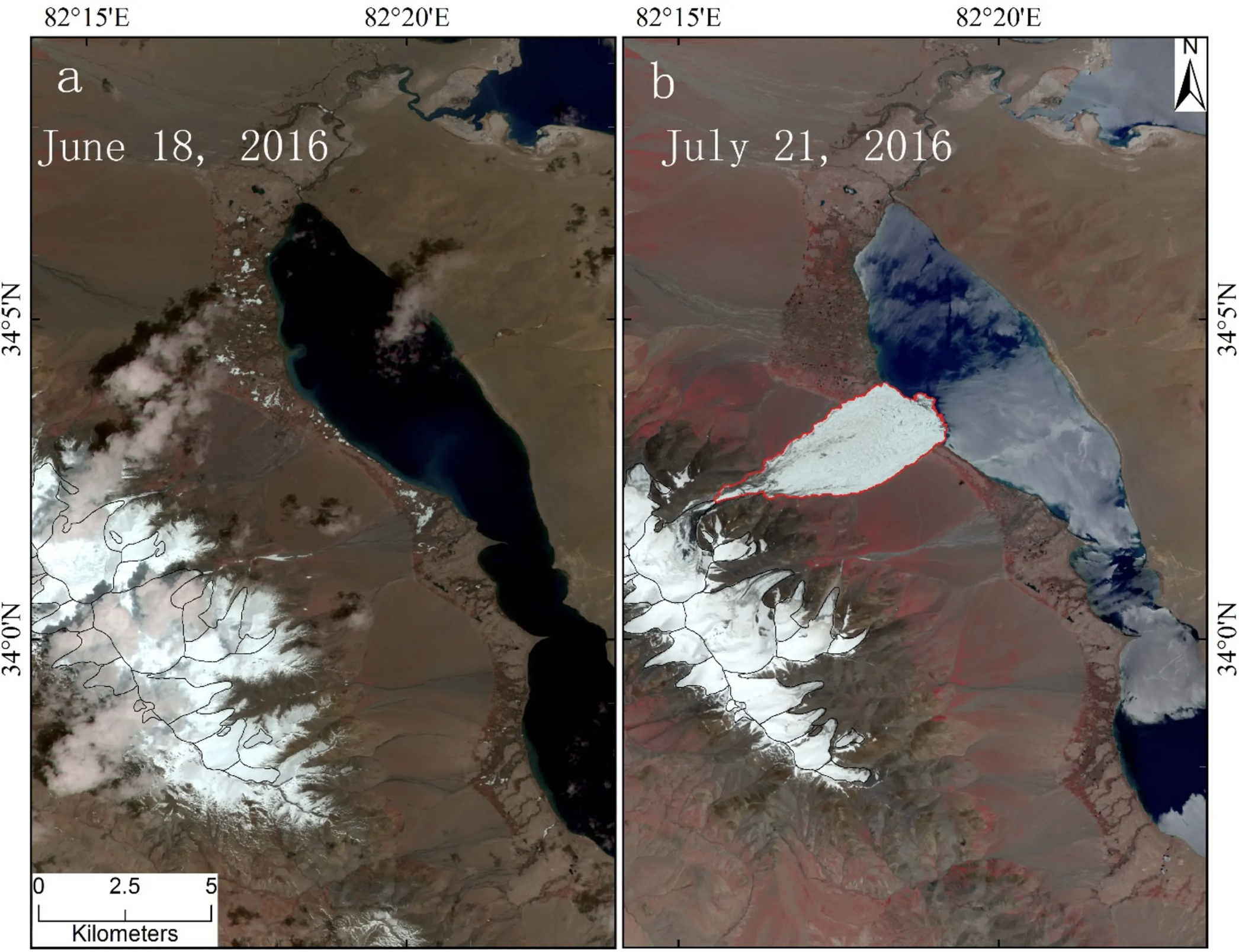}
    \caption{Images before and after the Aru Glacier collapse on 17 July 2016. Images are from Sentinel-2 on 18 June 2016 (a) and 21 July 2016 (b) with 10 m resolution. From \citep{tian2017twoglaciers}}
    \label{fig:aruavalanche}
\end{figure}

In an increasingly interconnected world, the ability to rapidly identify and respond to breaking events, particularly those that pose a threat to human life such as natural disasters or conflicts, is of paramount importance. The effectiveness of response efforts can be significantly impacted by the speed and accuracy of information gathering, analysis, and dissemination, with the potential to save lives and mitigate damage. However, obtaining such information poses significant challenges due to various factors. For example, there have been instances where news agencies faced difficulties in promptly or completely delivering information about ongoing events due to a scarcity of reporters present in the affected areas \cite{mumbaiattack}. 
The ubiquity of mobile devices and the near-constant network connectivity that characterizes our modern era have resulted in the generation of a massive amount of temporally- and spatially-stamped data (hereafter called mobile data). These datasets have been the subject of numerous studies aiming to derive individual human mobility patterns for various applications, including quantifying urban vitality \cite{sulis2018using}, modeling epidemic spread \cite{alessandretti_what_2022}, and inferring the activities of individuals \cite{liao2007extracting}. Simultaneously, the increasing number of orbital satellites has facilitated the collection of high-resolution images, capturing snapshots of geographical areas with sub-daily temporal frequency.
In this paper, we propose a novel data fusion methodology that integrates satellite imagery with mobile data to identify breaking events in real time. This approach leverages the strengths of both data types: mobile data provides an approximation of human mobility, proximity to one another, and the built environment, while satellite imagery offers visual information on physical changes to the built and natural environment.
This method holds significant potential for three critical applications. First, it can enhance the detection of small-scale disasters such as tornadoes, wildfires, and floods in rural regions, where sparse populations and limited infrastructure can impede timely detection and response. Second, it can augment search and rescue operations for lost hikers in remote wilderness areas, providing critical geospatial data to guide rescue teams and improve the chances of a successful rescue. Before and after images of the Aru Glacier avalanche in Figure \ref{fig:aruavalanche} suggest that large-scale changes in the natural environment can be inferred rather straightforwardly using satellite imagery \cite{tian2017twoglaciers}. Finally, it can aid in the identification of active conflict areas and population displacement in war-torn states, offering invaluable insights to humanitarian organizations and policy-makers. By leveraging the strengths of both satellite imagery and GPS traces, this approach overcomes the limitations of both data types and augments the ability to monitor, understand, and respond to dynamic events in diverse contexts. 
We demonstrate an application of the proposed method on a small-scale tornado that ripped through Muskogee, Oklahoma on May 15th, 2020. By combining mobile data from the area as well as satellite images before and after the event, we infer the occurrence of a natural disaster, identify damaged areas, and highlight hot spots for individuals' travel through an area (both spatially and temporally). 
\subsection{Our Contributions}
\begin{itemize}
    \item We develop a practice-oriented data fusion methodology to augment the inference task for a variety of use cases;
    \item We outline expected use cases and demonstrate the case study of a small-scale natural disaster for which the proposed method is appropriate, and;
    \item We provide an open-source implementation of the discussed methods.
\end{itemize}
The rest of the paper is organized as follows: Related Work discusses previous work on leveraging satellite images, GPS traces, social media posts, and the fusion of such datasets in a variety of contexts. Dataset and Software describes the datasets and the open-source software we employ. The Methodology section details our methods, which include data preprocessing steps as well as analysis specific to each data type. The Experiments section outlines our case study on a tornado in Muskogee, Oklahoma, while the Use Cases section describes other potential use cases for the proposed method. We conclude with a brief discussion of the implications of our findings.

\section{Related Work}
\label{sec:relatedwork}

Historically, capturing changes in the built and natural environment often involved on-the-ground data collection and analysis \cite{Haddad2011}. Surveys and field observations have been the most traditional method to capture such changes, and continue to be so today in many industries. Prior to the advent of modern satellites, aerial photography taken from aircrafts was another common method to document large-scale changes to the built environment \cite{Bewley2003}. While these two methods of data collection are sufficient for many studies, they (a) lack the immediacy called for in urgent real-time scenarios and (b) tend to miss developments in the absence of the proper equipment or manpower. In the last three decades, new methods of data collection in the built and natural environment have become prominent, including the use of satellite images, GPS traces from mobile devices, and user-generated content from social media platforms \cite{jongman2015early, ji2018identifying, yin2021multimodal}.

This literature review aims to provide a comprehensive understanding of current and recent research in the fields of mobile data analysis, satellite imagery, and crowdsourcing, highlighting works that have fused any two of them to assist in the inference task. Table \ref{tab:lit review} summarizes some of the papers discussed in the following paragraphs.

\begin{table}[!ht]
	\caption{Studies that have used mobile data/satellite imagery for various purposes.}\label{tab:studies}
	\begin{center}
		\begin{tabular}{l l p{4cm} p{6cm}}
			Author & Date & Datasets Used & Applications \\\hline
			Jongman et al. & 2015 & Satellite imagery, \newline Twitter posts &  Early flood detection in the Philippines and Pakistan\\
            Molinier et al. & 2016 & Satellite imagery, \newline mobile app data & Deriving biomass maps of forests \\
            Fayne et al. & 2017 & Satellite imagery & Flood detection in the Lower Mekong Basin \\
            Ji et al. & 2018 & Satellite imagery & Collapsed building detection in 2010 Haiti Earthquake \\
            Sulis et al. & 2018 & Transit smart card data, Twitter posts & Quantifying diversity and vitality in London \\
            Said et al. & 2019 & Satellite imagery, \newline social media posts & Natural disaster detection \\
            Pokhryiyal et al. & 2020 & Satellite imagery, \newline GPS traces & Estimation of poverty levels in Haiti \\
			Yin et al.  & 2021 & Satellite imagery, \newline GPS traces & Road attribute detection in Singapore and Jakarta \\
            Belcastro et al. & 2021 & Social media posts & Small-scale emergency detection
			\\\hline
		\end{tabular}
	\end{center}
\label{tab:lit review}
\end{table}

\subsection{Satellite Imagery and Remote Sensing}

The use of satellite imagery for event detection and inference has been well-documented. Satellite images provide a bird's-eye view of the Earth's surface, enabling the detection of physical changes over time. Though the granularity of satellite images alone may not provide the nuanced information needed in certain scenarios, they are useful in understanding big-picture changes on the built and natural environment. Particularly in the last five years, the proliferation of commercial satellite imagery providers has enabled access to this data type at a short notice and in real-time. This has been used in various applications, such as disaster management \cite{voigt2007satellite}, urban planning \cite{Albert2017, Karunanithi2016}, and environmental monitoring \cite{xu2010analysis}.

Recent advancements in machine learning have further enhanced the capabilities of satellite imagery analysis. For instance, convolutional neural networks have been used to automatically detect and classify objects in satellite images \cite{zhang2016deep}. Meta's novel object classification method, \textsc{Segment Anything} \cite{kirillov2023segany}, has been proposed for the geospatial domain with great promise \cite{qiusheng_wu_2023_7966658}. These developments have opened up new possibilities for event detection and inference, as it allows for the automated analysis of large volumes of satellite imagery and greatly reduces image processing time, allowing for subsequent immediate response on the ground. 

\subsection{GPS Traces from Mobile Devices}

Parallel to the advancements in satellite imagery analysis, the proliferation of GPS-enabled mobile devices has led to an explosion of mobile data, providing detailed information about the movements of individuals and groups \cite{zheng2008understanding}. Research has delved into the use of mobile data for tracking population movements \cite{bengtsson2011}, understanding social behavior \cite{Hong2008, calabrese2010real}, predicting and managing traffic patterns \cite{herrera2010evaluation}, and more.

The richness of this data source has also led to the development of new methods for event detection and inference. For instance, GPS traces have been used to detect social events \cite{Castro2014, Pan2013}, infer the activities of individuals \cite{liao2007extracting}, and understand where individuals conduct their daily activities \cite{chen2014traces}. However, the fusion of mobile data with other data sources for real-time event identification and inference in the built and natural environment, particularly in the context of emergencies and disasters, is an area that deserves further exploration.

\subsection{Crowdsourcing and Social Media Data} 

Crowdsourcing has emerged as a powerful tool for gathering and analyzing data, particularly in the context of event detection and response. The term "crowdsourcing" refers to the practice of obtaining information or input into a task or project by enlisting the services of a large number of people, either paid or unpaid, typically via the internet \cite{Estelles2012}. In the context of event detection, crowdsourcing can be used to gather real-time information from individuals who are directly experiencing or observing the event \cite{goodchild2007citizens}. One of the most notable examples of crowdsourcing in event detection is the Ushahidi platform, which was used to gather and map reports of violence in Kenya after the post-election violence in 2008 \cite{okolloh2009ushahidi}. Since then, Ushahidi has been used in various other contexts, including disaster response and environmental monitoring. 

Crowdsourcing is closely linked to social media's rise to prominence as a crucial news outlet during disasters. A mountain of literature has been published on the promises of social media data in augmenting the inference task in a range of scenarios. Social media websites provide a platform for individuals to share real-time information on developing events, including their impacts and the needs of affected communities \cite{palen2007crisis}. The information circulated on social media tends to dissipate quicker and be more specific than official news outlets, as anyone and everyone has the ability to share and post \cite{Palen2008}. This information has been used by disaster response organizations to better understand the situation on the ground and coordinate their response efforts.

Several studies have explored the use of social media data in disaster response. For instance, \citep{sakaki2010earthquake} developed a system that uses Twitter data to detect earthquakes in real-time. Similarly, \citep{Ashktorab2014TweedrMT} developed a system that uses Twitter data to extract actionable information during disasters, such as requests for help or reports of infrastructure damage. \citep{guan2014using} developed a metric to quantify the evolution of disasters based on Twitter activities, demonstrating its use for damage detection during Hurricane Sandy, affecting much of the eastern United States and various Caribbean nations. More recently, social media data has been combined with other data sources for disaster response. For example, \citep{jongman2015early} proposed a method for fusing social media data and satellite imagery to detect floods. This approach leverages the strengths of both data sources, using social media data to provide real-time information about the flood and satellite imagery to provide a macro view of the flood's impacts. We refer the reader to \citep{said2019natural} for a broader review of these approaches.

Despite the significant potential of crowdsourcing and social media data for event detection and response, there are notable limitations that must be acknowledged. The quality and reliability of crowdsourced data can vary significantly. Misinformation or inaccurate reports can be disseminated, either unintentionally due to confusion or panic or intentionally as a form of manipulation or cyberattack \cite{allcott2019trends}. This can lead to false positives in event detection or misdirected response efforts. Additionally, crowdsourcing is inherently dependent on the availability and willingness of individuals to contribute information. In certain scenarios, such as in remote wilderness areas or war-torn regions, the number of individuals able to contribute data may be limited. Similarly, during small-scale or rapidly unfolding events, there may not be sufficient time for crowdsourced information to be gathered and analyzed. These limitations highlight the need for complementary approaches to event detection and response. Our proposed method addresses these limitations by providing a reliable, real-time source of information that is not dependent on individual contributions or vulnerable to the same misinformation risks. 

\subsection{Data Fusion for Event Inference}
While there is a substantial body of research on the use of mobile data and satellite imagery independently, there is a noticeable gap in the literature when it comes to their combined use for real-time event identification and response. This gap represents a significant opportunity for new research that can contribute to both academic knowledge and practical applications in disaster management, conflict monitoring, as well as search and rescue operations.

The fusion of satellite imagery and GPS traces for event inference is a relatively new area of research and is uniquely poised to overcome the limitations placed on only analyzing each data source individually. Though GPS traces have emerged as a prominent data source, it should be noted that such datasets tend to exclude certain populations, such as the extreme poor \cite{Pokhryiyal2020}. In developing countries, this population may represent a significant share of the population. Thus, in the event of an emergency, mobile data cannot be the sole indicator of irregular movement patterns. Satellite imagery itself may also be inadequate in event inference due to the lack of specific mobility information. However, when combined, these two data sources can supplement each other and leverage the additional information to better inform the inference task.

Though previous studies have explored the intersection of these two data sources, the direction of application we intend to use our tool differs from previous works. For instance, \citep{yin2021multimodal} proposed a method for fusing satellite imagery and GPS traces to infer and fill in road attributes such as speed limits and lane widths, with the goal of developing a more comprehensive view of road networks globally. \citep{Pokhryiyal2020} fused satellite images with mobile data to estimate poverty levels in Haiti, where a nationwide census had not been conducted since 2012. In developing countries where there may not be the economic resources necessary to complete annual censuses, using a combination of data sources to estimate various baselines may emerge as a crucial tool for financial aid allocation and future urban planning. A combination of mobile data and satellite imagery has also been used to study forests and obtain biomass measurements \cite{Molinier2016}. This study combined citizen science-powered mobile data collection and satellite imagery to garner forest information and produce biomass maps in a cost effective manner. 

Notably, both the work of \citeauthor{yin2021multimodal} and \citeauthor{Pokhryiyal2020} provide information that can be used for future planning. However, we hope to use GPS traces and satellite imagery in real time for instantaneous data analysis and event identification. Additionally, unlike in \citeauthor{Molinier2016}, where mobile data is actively collected through an app that must be user-operated, we obtain GPS traces passively through users' interactions with various apps (not a specific one). Thus, our event detection and inference capabilities are completely powered by remote data, and without the need for active, real-time human participation.

\section{Dataset and Software}
\label{sec:dataset}

For mobile data, we employ privacy-protected, passively generated GPS data from an American data solution provider specializing in geospatial analytics. The dataset contains discrete GPS points of 2,000 anonymous, opted-in individuals in the Tulsa metropolitan area between December 2019 and July 2020. The locational data recorded includes timestamps, unique device identifiers, latitudes and longitudes, and a measure of data precision (i.e., a spatial radius for which the provider has 95\% confidence in the reported coordinates). In addition to anonymizing the data, the data provider obfuscates home locations to the census-block group level, and removes sensitive points of interest from the dataset, in order to preserve privacy. It's also crucial to note that the temporal granularity of this data is determined by user activity, meaning data points are generated irregularly and not at fixed intervals. This non-uniformity of temporal data represents the actual, inconsistent frequency of device usage among the sampled population.

For satellite imagery, we pull data from Planet Labs, a private American company that specializes in Earth imaging. Planet operates a large constellation of miniature satellites, known as Doves, which continually capture and transmit images of Earth. The 3-band images we use have undergone orthorectification, a process that corrects the geometric distortions in an image to represent a flat surface, thereby ensuring that the scale is uniform throughout. This makes the images reliable for distance measurements and for overlaying with other geospatial data. Additionally, the images have been color-corrected to ensure that the colors represented are as close as possible to what the human eye would perceive if viewing the scene directly. This enhances the clarity and interpretability of the images, making them a valuable resource for our research.

We implemented our methodology in the \textsc{Python} programming language \cite{CS-R9526}. We used \textsc{pandas} \cite{mckinney-proc-scipy-2010} to read and wrangle mobile data and \textsc{Rasterio} \cite{Rasterio} to manipulate satellite images. We also leveraged \textsc{NumPy} \cite{harris2020array} for a variety of array operations, \textsc{GeoPandas} \cite{geopandas} for operations involving coordinates, and \textsc{scikit-mobility} \cite{pappalardo2019scikit} to conduct mobile data cleaning and preprocessing. Last but not least, we used \textsc{matplotlib} \cite{Hunter:2007} to produce any visualizations of our results. Any other chunk of code we developed can be found on our GitHub repository: \url{https://github.com/ekinugurel/SatMobFusion}.

\section{Methodology}
\label{sec:methodology}
\begin{figure}[!h]
  \centering
  \includegraphics[width=0.8\textwidth]{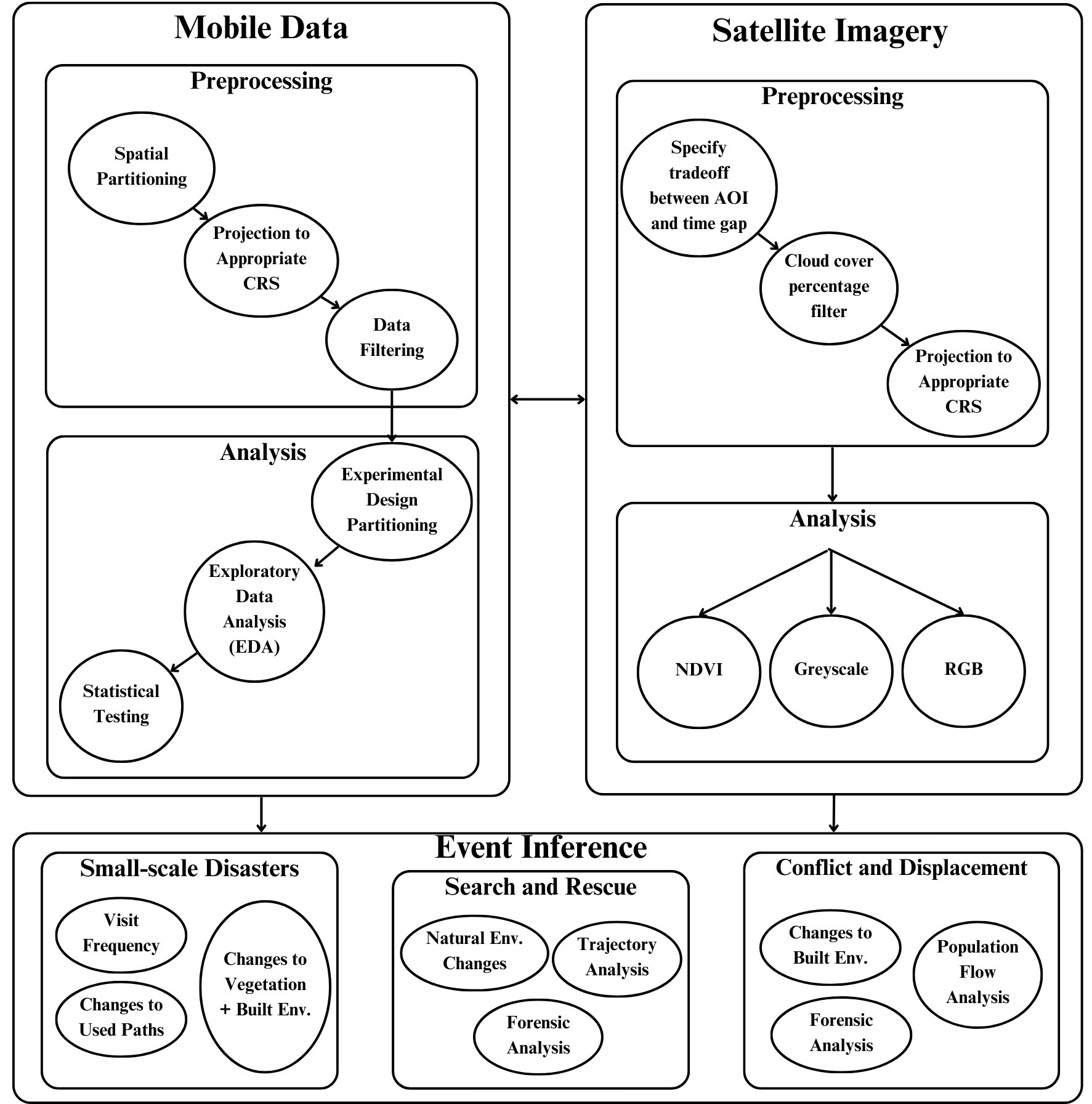}
  \caption{Methodological flowchart; the top bins describe the order of processes employed in this paper, while the bottom outlines key geospatial analysis metrics to carry out the inference task}
  \label{fig:method_flowchart}
\end{figure}

Our methodology has two modules---one that deals with the analysis of mobile data, and another with that of satellite imagery. In both modules, we discuss relevant preprocessing methods as well as analysis steps to aid in the inference task. Figure \ref{fig:method_flowchart} shows the high-level steps in preprocessing and analysis of both datasets while also outlining key geospatial analysis processes for the three use cases discussed in the Introduction.

\subsection{Mobile Data}
For mobile data, the general preprocessing workflow is as follows:

\begin{enumerate}
  \item Spatial Partitioning
  \item Coordinate Reference System (CRS) Projection
  \item Data Filtering
\end{enumerate}

To avoid wastefulness in computation, we begin by partitioning the mobile data spatially, which we do by drawing a bounding box using four sets of coordinates. In practice, there are several ways to achieve this at scale: Leveraging a geocoding application protocol interface (API) to obtain the coordinates based on an existing point of interest, partitioning by federal and/or state designations such as census tracts, ZIP codes, or county borders, manually typing in a bounding box (i.e., by defining the geographic coordinates of the corners of the box), or obtaining a KMZ file from the 
National Oceanic and Atmospheric Administration's \href{https://apps.dat.noaa.gov/stormdamage/damageviewer/}{Damage Assessment Toolkit}. For this effort, we used the last option to derive the centroid coordinates of the affected area around which we drew a $1 \;\textrm{km}^2$ bounding square.

Once the data is spatially partitioned, we project coordinates to the appropriate CRS. Different CRSs are designed to accurately represent different aspects of the Earth's surface. Some are designed to preserve area, while others preserve shape, distance, or direction. In the context of event inference, the optimal choice of CRS should minimize bias in the derived mobility metrics, which may exist due to distortions of distances and areas. Therefore, we use an equal-area projection around UTM Zone 15, which covers the area around our case study.

We then filter out any erroneous data points based on a threshold of segment velocity. Since this is done "as the crow flies" and not with respect to the underlying transportation network, this threshold should be set conservatively. Velocity filtering discards infeasible jumps and oscillations in trajectories, which may occur as a result of tall buildings blocking a satellite's line of sight to the mobile device (termed "urban canyon effect", \cite{ben-moshe_improving_2011}). These oscillations may also occur due to low data precision as a result of the "cold-start problem"---when a signal dropped by WiFi is not immediately picked up by a satellite, requiring some time before being able to acquire navigation data and calculate a position \cite{lehtinenTTFF}. 

Once preprocessing is finished, we conduct inference-related analyses using the following steps:
\begin{enumerate}
    \item Experimental Design Partitioning
    \item Exploratory Data Analysis (EDA)
    \item Statistical Testing
\end{enumerate}

Specifically, we first define pre-, during-, and post-event periods that temporally segment the data. The purpose of this partitioning is to derive key metrics for each period, which would assist in detecting anomalies and enable more formal statistical tests. The duration of these periods depends on the nature of the analysis---for natural disasters like tornadoes or hurricanes, the "during" period may be a few hours, while for applications in search \& rescue or in conflict areas, the same period may only last a few minutes. 

In conducting EDA on time-series mobile data, the objectives are to understand the structure of the data (i.e., identifying immediate trends, patterns, or outliers), detect seasonality (i.e., periodic fluctuations), and prepare the data for more formal anomaly detection techniques and statistical tests. The EDA process can also assist in formulating hypotheses about possible causes for anomalies by exploring relationships between the target variable and other variables. In the context of mobile data within constrained geography, the metrics we analyze include the radius of gyration $r_g$ (the characteristic distance traveled by an individual in a given period, Equation \ref{eq:rog}) \cite{song2010limits}, the number of extended stays at a particular location $c$ (Equation \ref{eq:numberofstays}), and the number of visits per time unit $v$.
\begin{equation}
    r_g = \sqrt{\frac{1}{n} \sum_{i=1}^n (r_i - r_{cm})^2}
\label{eq:rog}
\end{equation}
\noindent where $r_i$ represents the $i = 1, \ldots, n$ locations recorded for the individual and $r_{cm} = \frac{1}{n} \sum_{i=1}^n r_i$ is the center of mass of the period's trajectory. 
\begin{align}
    c &= \sum_{i=1}^n c_i, \\
    c_i &= \begin{cases}
        1 & \text{if } t_i = 1, l_i = 1\\
        0 & \text{otherwise}
    \end{cases}
    \label{eq:numberofstays}
\end{align}
\noindent where $t_i$ is an indicator variable denoting whether a device is observed more than once within a 15-minute period and $l_i$ is an indicator variable denoting whether multiple observations of the device are within a 100-meter radius (only considered if $t_i = 1$). We note that these thresholds are robust and suggest adjusting them to suit the underlying domain. 
\begin{equation}
    v = \sum_{j=1}^m v_j
    \label{eq:visitspertime}
\end{equation}
\noindent where $j = 1, \ldots, m$ indexes the subsections of the inspected geographical area and $v_j$ is an indicator variable denoting whether the individual was located in that subsection in the given period.

Finally, we conduct statistical tests to confirm the presence of anomalies. In this paper, we employ the $Z$-score method (Equation \ref{eq:zscore}), in which the principle is to assume the data follows a normal distribution and then to identify data points that are too far from the mean.
\begin{equation}
    Z = \frac{x - \mu}{\sigma}
\label{eq:zscore}
\end{equation}
\noindent where $x$ is a data point, $\mu$ is the mean of the dataset, and $\sigma$ is the standard deviation. We consider Z-scores above 3 (or below -3) an anomaly.

\subsection{Satellite Imagery}

The general workflow for processing and analyzing satellite images is as follows: 
\begin{enumerate}
    \item Specify tradeoff between region-of-interest (ROI) coverage and time gap to the event of interest
    \item Pull imagery and project to CRS
    \item Image Analysis: RGB, greyscale, and NDVI differences
\end{enumerate}

Providers like Planet Labs have multiple orbital satellites whose coverage area may intersect---the goal of each satellite is to pass over a particular geography at least once per day; however, various factors can influence the availability of imagery and cause irregular gaps in observation, including cloud cover, technical issues with onboard cameras or the image transmission process, and regulatory restrictions imposed by nation-states. For these reasons, we derive a utility metric (Equation \ref{eq:coveragetradeoff}) to determine the best available before and after image for an event of interest. This metric weighs the image's area of coverage against its time gap with the event using a tradeoff coefficient---the idea is to maximize the spatial overlap between the image and the ROI while minimizing the number of days between the images. 
\begin{equation}
\label{eq:coveragetradeoff}
    u = \frac{A_{img}}{A_{evt}} - \frac{|t_{img} - t_{evt}|}{\phi}
\end{equation}
\noindent where the first term denotes the percent of ROI covered by the image (i.e., $A_{img}$ is the area of the image and $A_{evt}$ is the area affected by the event), the numerator of the second term denotes the number of days between the images, and $\phi$ is the coverage tradeoff coefficient. In this work, we use $\phi = 0.25$, meaning that 25\% (percentage points) more coverage of the AOI is defined to be worth as much as the image being taken 1 day before or after the event. 

As an additional filter to the above utility metric, we require all images to have less than 50\% cloud coverage. Once we have the satellite images, we project them to the same CRS we used for the mobile data to avoid misalignment. 

We analyze the images to identify any physical changes to the built and natural environment. This analysis involves looking at the regular visible spectrum (RGB), a grayscale version (i.e., luminosity, Equation \ref{eq:greyscale}), and the Normalized Difference Vegetation Index (NDVI, Equation \ref{eq:NDVI}), which measures the greenness and the density of the vegetation captured in a satellite image.
\begin{align}
\label{eq:greyscale}
    NDVI &= \frac{NIR - R}{NIR + R} \\
    GS &= 0.299 R + 0.587 G + 0.114 B
\label{eq:NDVI}
\end{align}
\noindent where $NIR$, $R$, $G$, and $B$ are near-infrared, red, green, and blue reflectances, respectively. NDVI is always between -1 and 1, where negative values signify the presence of clouds, water, or snow, values close to zero signify earthy rocks and bare soil, and positive values signify the existence of some vegetation (with more positive meaning more green).

\begin{figure}[!h]
  \centering
  \includegraphics[width=0.8\textwidth]{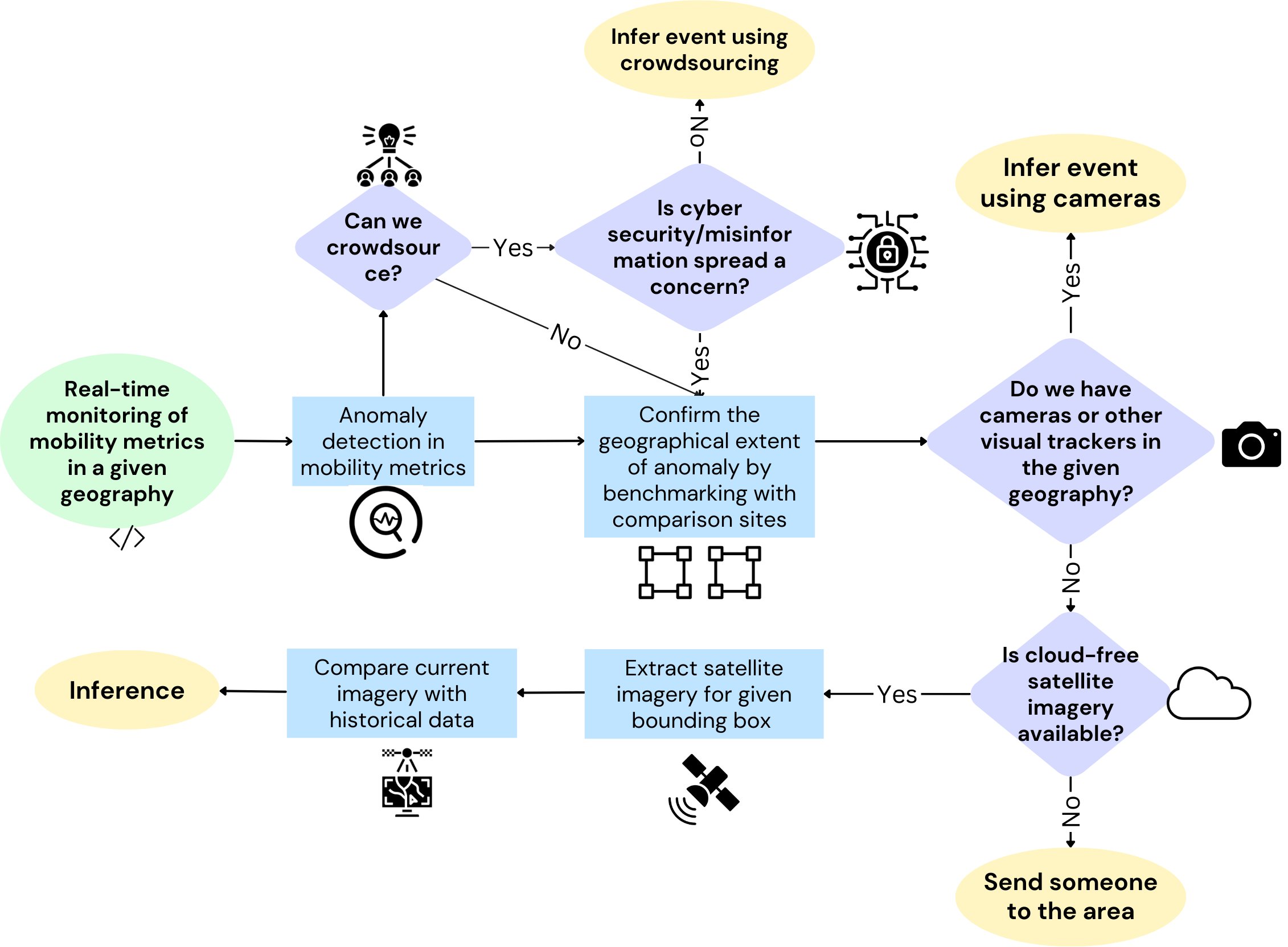}
  \caption{Usage decision tree for the proposed method}
  \label{fig:framework}
\end{figure}

Figure \ref{fig:framework} shows a decision tree outlining when the proposed method should be employed based on a series of conditions, organized by the cognitive load required to succeed in the inference task. Intuitively, crowdsourcing may be the preferred option for most organizations and individuals but can be unreliable due to the threat of misinformation or cybersecurity breaches. Similarly, while cameras and other in-situ sensors could provide longer visual feeds that are also easier to interpret, their availability in certain settings tends to be sparse. If the above two conditions are true, and cloud-free satellite imagery is available, then employing our method may be of interest to organizations and individuals alike.

Relating to the above figure, we emphasize three points about the proposed methodology: 
\begin{enumerate}
    \item The process of event detection starts with mobile data, not satellite imagery, due to the feasibility of passively monitoring geography based on mobility metrics. However, the fusion of the two sources is what enables us to make an inference.
    \item The proposed methodology is designed to be flexible, allowing the user to define an event of interest using a configuration file.
    \item This approach can be used with both real-time and historical data, depending on the context of the application, making it a potentially powerful tool for improving emergency response times and information percolation speeds.
\end{enumerate} 

\section{Case Study}
\label{sec:experiments}

To validate our proposed methodology, we conducted a case study on the EF-1 tornado that hit near Muskogee, Oklahoma on May 15, 2020. This event provided an opportunity to test our approach in a real-world scenario involving a small-scale disaster in a rural region.

\begin{figure}[!]
\centering
\begin{subfigure}{0.49\textwidth}
     \includegraphics[width=\textwidth]{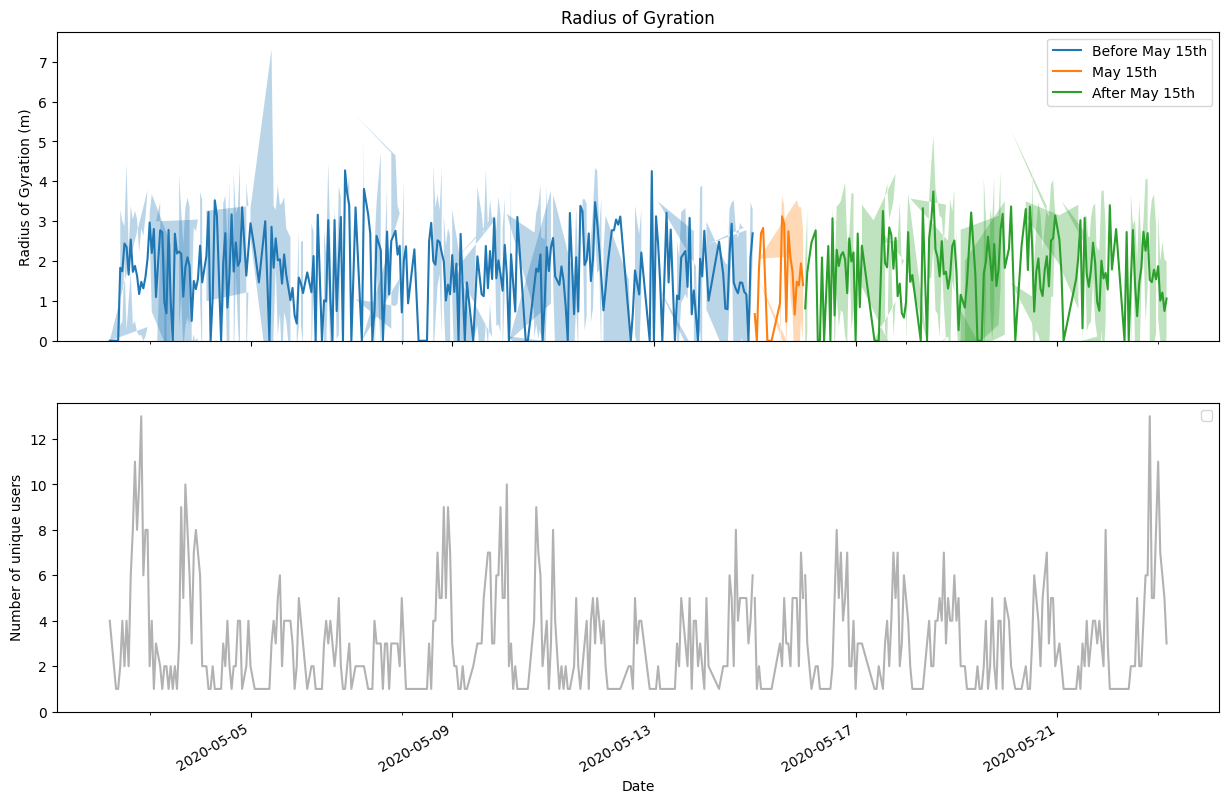}
     \caption{Radius of Gyration}
     \label{fig:rog}
 \end{subfigure}
\hfill
\begin{subfigure}{0.49\textwidth}
    \includegraphics[width=\textwidth]{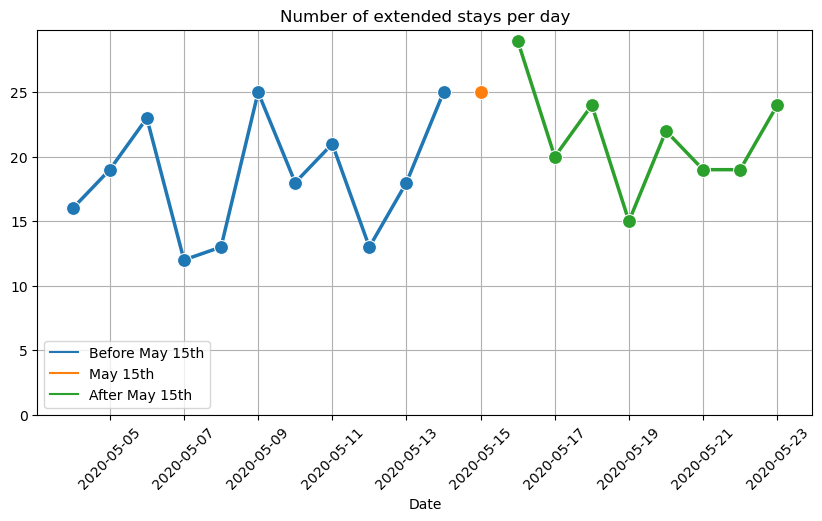}
    \caption{Number of Stays per day}
    \label{fig:stays}
\end{subfigure}
\caption{EDA of mobility metrics in the before, during, and after periods}
\label{fig:EDA}
\end{figure}

After spatially partitioning the GPS traces to include the subset within the bounding box of the tornado, we projected all data to UTM Zone 15 and filtered the data by the logic described in Section \ref{sec:methodology}. Then, we defined the two weeks before the day of the event (May 2 - May 15) as the before, the day of the event (May 15) as the during, and the week after the event (May 15 - May 22) as the after periods and began exploring the data. Figure \ref{fig:EDA} visualizes the time series of two metrics, highlighting the periods by color; there are no noticeable changes in the radius of gyration, while the number of extended stays at a given location within the ROI increases sharply the day after the event. However, Z-score tests on both metrics do not detect any anomalies.

\begin{figure}[!]
    \centering
    \includegraphics[width=1.0\textwidth]{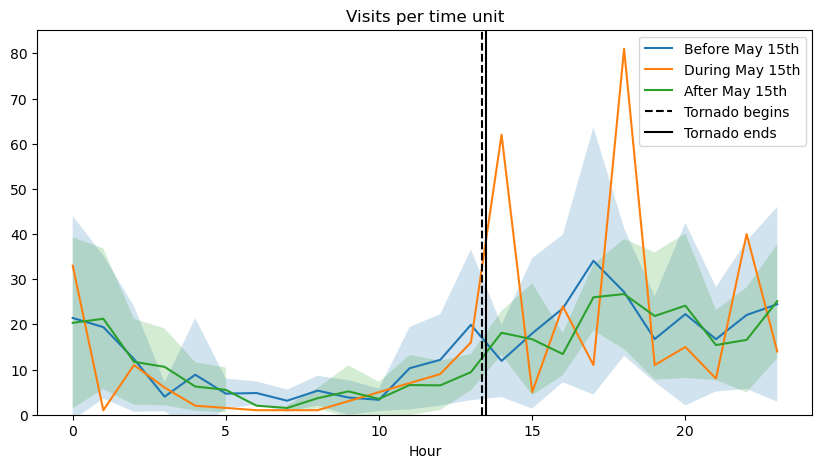}
    \caption{Visits per hour to the ROI in the before, during, and after periods sorted by hour of day}
    \label{fig:visits}
\end{figure}

However, when we look at the third metric (visits per hour to the ROI, Figure \ref{fig:visits}), we see an unmistakable outlier; on the day of the tornado, visits to the area increased sharply at 2 pm and again at 6 pm, substantially beyond the usual fluctuations captured by the 95\% confidence interval (light blue shade) in the before and after periods. Furthermore, Z-score tests on this metric confirm the hypothesis: Two anomalies exist in the data and they are at 2 pm ($Z = 3.47$) and at 6 pm ($Z = 4.87$) on May 15th. 

Following the anomaly detection, we pull in satellite imagery as described in the previous section. We use the utility metric in Equation \ref{eq:coveragetradeoff} to land on two images, one from May 9 and another from May 18. These two images are free of clouds and strike a balance between the percent of ROI covered and the number of days separating them from the event.

\begin{figure}
\centering
\begin{subfigure}{0.49\textwidth}
     \includegraphics[width=\textwidth]{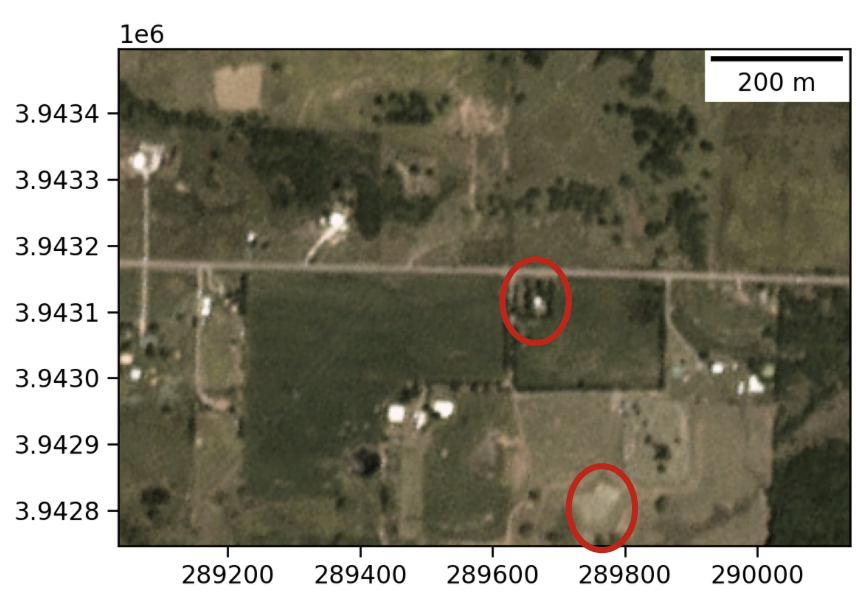}
     \caption{Before the Muskogee tornado on May 9, 2020}
     \label{fig:beforeRGB}
 \end{subfigure}
\hfill
\begin{subfigure}{0.49\textwidth}
    \includegraphics[width=\textwidth]{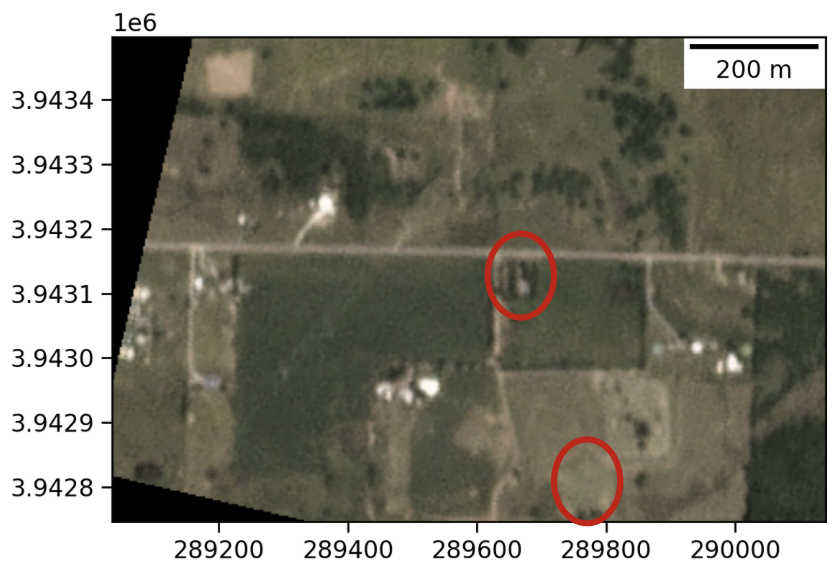}
    \caption{After the Muskogee tornado on May 18, 2020}
    \label{fig:afterRGB}
\end{subfigure}
\hfill
\begin{subfigure}{0.49\textwidth}
    \includegraphics[width=\textwidth]{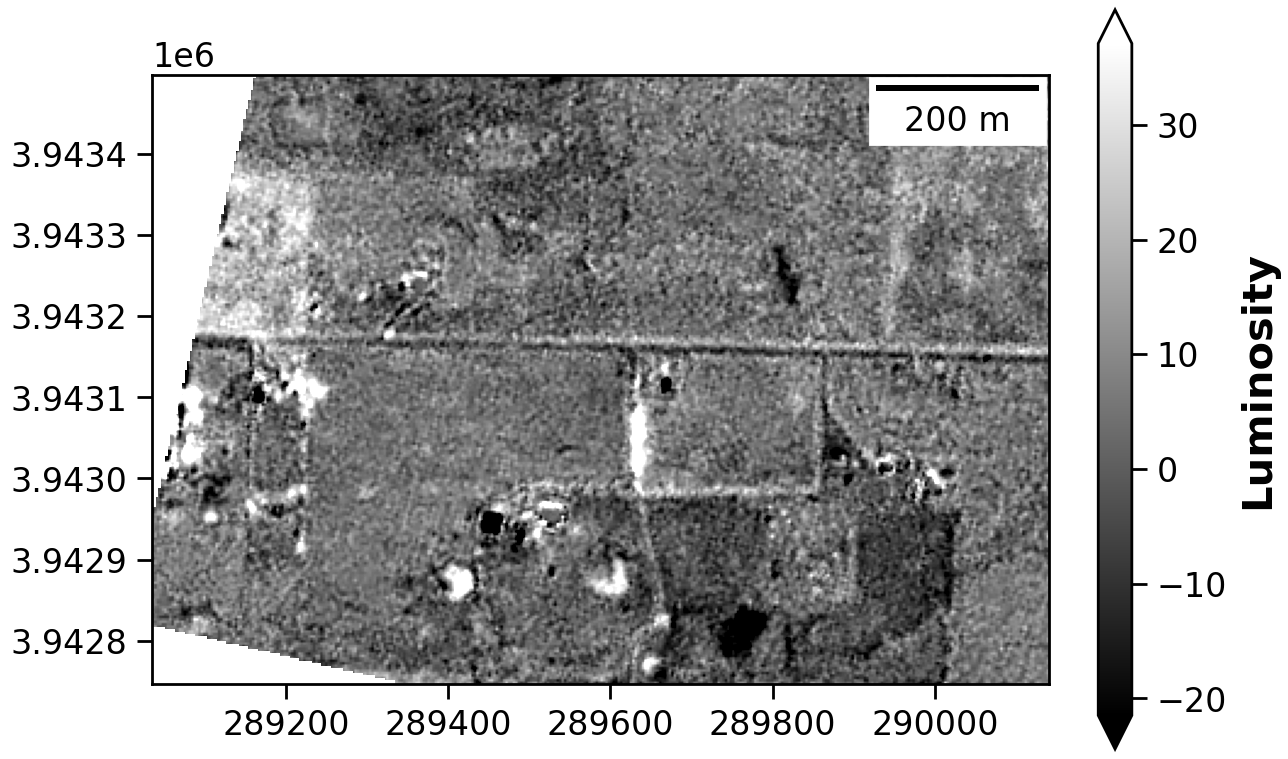}
    \caption{Greyscale difference between the two images}
    \label{fig:GSdiff}
\end{subfigure}
\hfill
\begin{subfigure}{0.49\textwidth}
    \includegraphics[width=\textwidth]{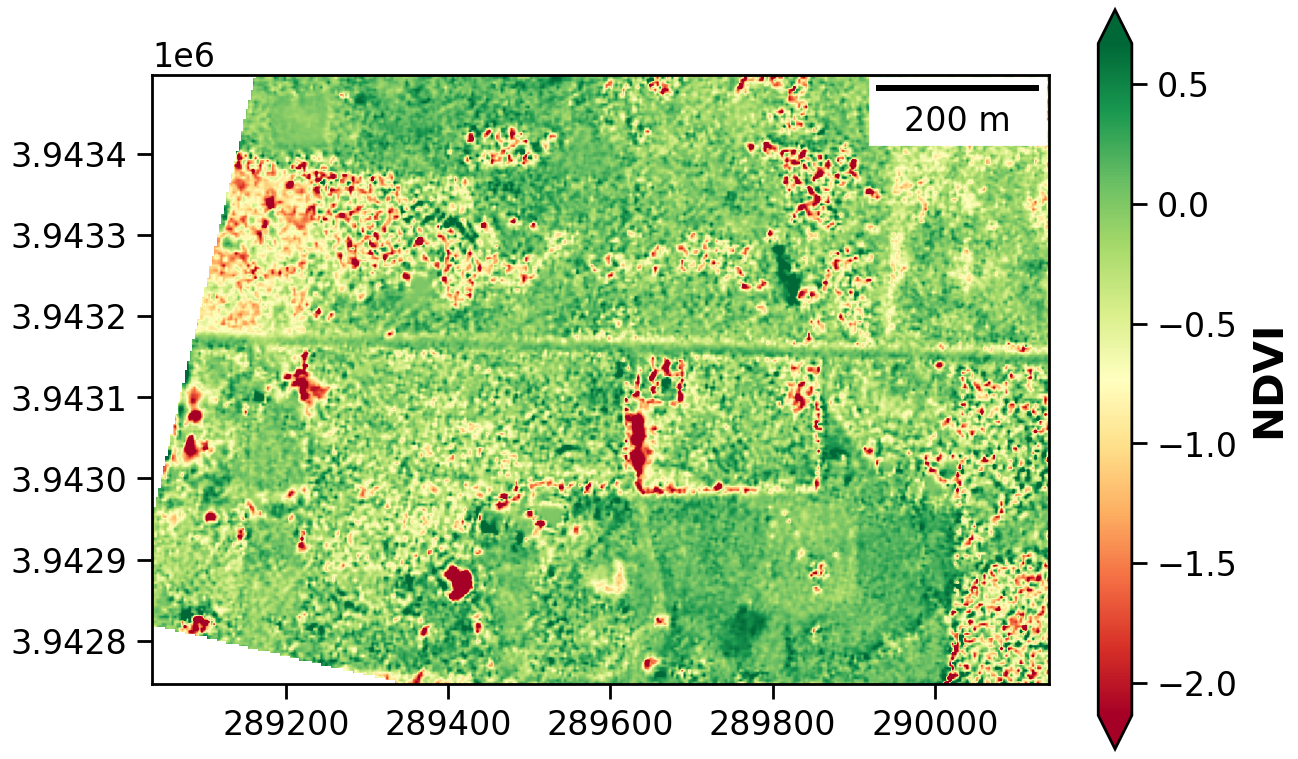}
    \caption{NDVI difference between the two images}
    \label{fig:NDVIdiff}
\end{subfigure}
\caption{Satellite imagery with 10m resolution documenting changes to the built and natural environment after the Muskogee tornado}
\label{fig:satimg}
\end{figure}

We analyze the images based on visual inspection, the greyscale difference, and the NDVI difference (Figure \ref{fig:satimg}). We highlight two spots on the regular color band images in which changes to the soil and a corner building are apparent to the naked eye. Furthermore, while noisy, the greyscale difference image shows changes to luminosity and the NDVI difference image shows loss of vegetation throughout the captured area (note that 0.0 is shown in light green).

The images show clear evidence of physical changes to the ROI. When combined with insights from GPS traces, and even without hindsight bias, we would infer that a mildly-destructive natural disaster occurred, piquing passerby's interest as well as landowners' concern on the day of the event (resulting in greater visits to the ROI).

\section{Use Cases}
\label{expectedusecases}
The proposed methodology is expected to outperform existing methods for data collection and information percolation for a range of use cases. We especially anticipate broad applicability in the following settings: 

\begin{itemize}
    \item \textbf{Intelligent transportation systems}:
    The likelihood of finding parking approximate to a desired location and in a timely manner has been shown to be a determinant of travelers' mode choice \cite{assemi2020searching}. Therefore, providing accurate parking-related information can augment an individual's decision-making process. Public and private entities can leverage the proposed method to provide real-time information on parking availability. The temporal sparsity of satellite imagery can be augmented with GPS traces to provide an online learning framework for the likelihood of finding parking.
    \item \textbf{Events in rural areas}:
    In rural areas where traditional data collection methods may be sparse or non-existent, our methodology could provide valuable insights into events as they unfold. The combination of satellite imagery and mobile data could help identify and respond to emergencies such as wildfires, floods, or lost hikers. 
    \item \textbf{Extreme-weather events}: 
    During extreme-weather events like hurricanes, tornadoes, or blizzards, our methodology could provide real-time information about the event's impact and aid in coordinating response efforts. The ability to quickly identify areas of high impact could help prioritize resource allocation and rescue operations.
    \item \textbf{War-torn states}:
    In regions affected by conflict, obtaining reliable on-the-ground information can be challenging. Satellite imagery combined with mobile data could help identify areas of active conflict, population displacement, or infrastructure damage.
    \item \textbf{Areas without network connectivity}: 
    In areas where network connectivity is limited or non-existent, such as remote wilderness areas or during large-scale network outages, satellite imagery can still provide valuable data. When combined with historical or intermittent mobile data, this could help identify and respond to events such as lost hikers or the spread of wildfires.
    \item \textbf{National security matters}: 
    Our methodology could also be used in the context of national security, helping to identify potential threats or unusual activities. The fusion of satellite imagery with mobile data could provide a more comprehensive picture of activities across a wide geographical area.
\end{itemize}

In each of these use cases, our methodology aims to provide a more nuanced and timely understanding of events as they unfold, potentially improving response times and outcomes.

\section{Conclusion}
\label{sec:conclusion}
We have presented a novel data fusion methodology combining satellite imagery with GPS traces generated from mobile devices. Our approach leverages the strengths of both data types: mobile data provides an approximation of human mobility, proximity to one another, and the built environment, while satellite imagery offers visual insights into physical changes to the built and natural environment. Furthermore, we have demonstrated the case study of a small-scale tornado, showing the promise of our method in identifying anomalies in movement patterns using mobile data, changes to the built environment using satellite imagery, and inferring the cause of both using our fusion methodology. 

Our methodology is viable due to the increasing accessibility of commercial data providers specializing in remote sensing and mobile devices. This trend is only expected to continue---the number of active orbital satellites has grown exponentially since 2017 \cite{orbi_sat}, while the percentage of Americans that own a smartphone has risen to 97\% \cite{mobile_device_usage}. The continued commercialization of such products will enable researchers and practitioners to monitor, categorize, and understand the natural and built environments in ways unimaginable just 50 years ago. We hope our method is a step in this direction, enabling others to pursue endeavors made possible by the fusion of these data sources.

\section{Data Access Statement}
The authors note that only Ekin Ugurel had access to the privacy-protected GPS dataset. Data supporting this study cannot be made available due to privacy-preserving research agreements with the data provider.

\section{Author Contributions}
The authors confirm contribution to the paper as follows: study conception and design: E. Ugurel, S. Coenen; data collection: E. Ugurel, S. Coenen; analysis and interpretation of results: E. Ugurel, S. Coenen; draft manuscript preparation: E. Ugurel, M. Chen, S. Coenen, C. Chen. All authors reviewed the results and approved the final version of the manuscript.

\newpage
\bibliographystyle{abbrvnat}
\bibliography{trb_template}
\end{document}